\documentclass{article}
\usepackage{spconf,amsmath,graphicx, adjustbox, inputenc}
\usepackage{array}
\usepackage{geometry}
\usepackage{booktabs}
\usepackage{longtable}
\usepackage{tikz}
\usepackage{multirow}
\def\checkmark{\tikz\fill[scale=0.4](0,.35) -- (.25,0) -- (1,.7) -- (.25,.15) -- cycle;} 

\title{Multi-task Learning for Camera Calibration}
\name{Talha Hanif Butt, Murtaza Taj}
\address{Computer Vision and Graphics Lab\\
 Lahore University of Management Sciences, Lahore, Pakistan\\
 thanifbutt@gmail.com,  murtaza.taj@lums.edu.pk\\
 https://cvlab.lums.edu.pk/}
\begin{document}
%
\maketitle
\begin{abstract}
For a number of tasks, such as 3D reconstruction, robotic interface, autonomous driving, etc., camera calibration is essential. In this study, we present a unique method for predicting intrinsic (principal point offset and focal length) and extrinsic (baseline, pitch, and translation) properties from a pair of images. We suggested a novel method where camera model equations are represented as a neural network in a multi-task learning framework, in contrast to existing methods, which build a comprehensive solution. By reconstructing the 3D points using a camera model neural network and then using the loss in reconstruction to obtain the camera specifications, this innovative camera projection loss (CPL) method allows us that the desired parameters should be estimated. As far as we are aware, our approach is the first one that uses an approach to multi-task learning that includes mathematical formulas in a framework for learning to estimate camera parameters to predict both the extrinsic and intrinsic parameters jointly. Additionally, we provided a new dataset named as CVGL Camera Calibration Dataset~\cite{butt2022camera} which has been collected using the CARLA Simulator~\cite{Dosovitskiy17}. Actually, we show that our suggested strategy outperforms both  conventional methods and methods based on deep learning  on 6 out of 10 parameters that were assessed using both real and synthetic data. Our code and generated dataset are available at https://github.com/thanif/Camera-Calibration-through-Camera-Projection-Loss.

\end{abstract}
\section{Introduction}
\label{sec:intro}

Camera calibration is the process of identifying the six extrinsic (rotation, translation) and five intrinsic (focal length, image sensor format, and principal point)  characteristics of a certain camera. 


\begin{figure}
\centering
    {\includegraphics[width=0.95\columnwidth, trim={0 8cm 8cm 0},clip]{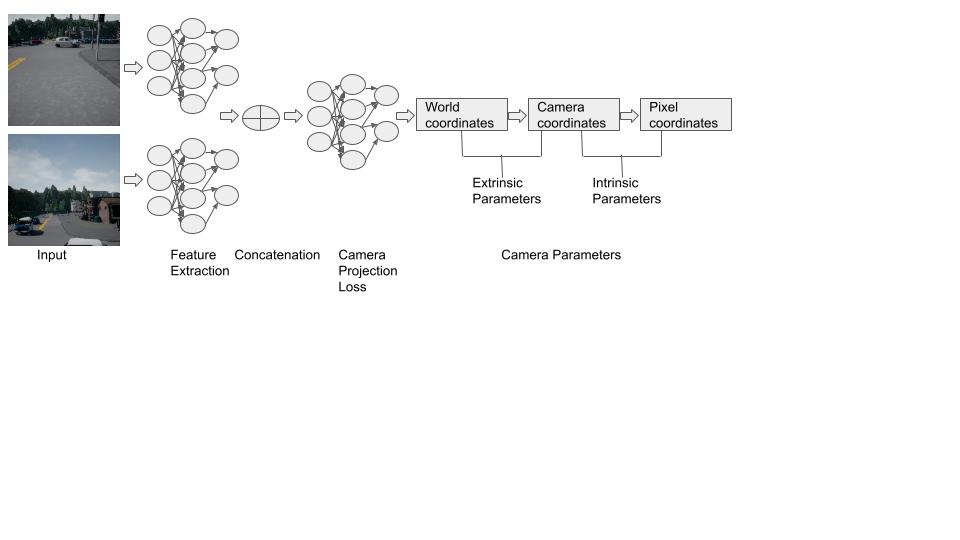}} 
    \caption{Using a trained version of the Inception-v3~\cite{szegedy2016rethinking} algorithm and the proposed Camera Projection Loss, our method calculates intrinsic (principal point offset) and extrinsic (baseline, pitch, and translation) parameters} 
    \label{fig:flowdiagram}
\end{figure}

\begin{figure}[t]
\begin{center}
\begin{tabular}{cc}
	\includegraphics[width=0.45\columnwidth]{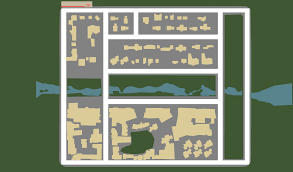}&
	\includegraphics[width=0.45\columnwidth]{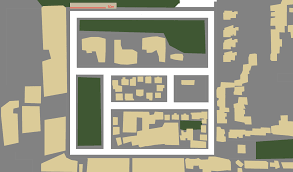}\\
\end{tabular}
\end{center}
\caption{Street maps of towns from CARLA Simulator (a) First Town. (b) Second Town.}
\label{fig:town}
\end{figure}

\begin{table*}[t]

\caption{Overview of some recent configurations for different aspects of Camera Calibration}
\centering
{
\begin{adjustbox}{max width=\textwidth}
\begin{tabular}{ |l|c|c|c|c|c|c|c|c|c|c|c| } 
\hline
 Paper&Input& Tilt & Roll & Focal Length & Radial Distortion & Pan & Rotation & Translation & Fundamental Matrix & Position & Orientation\\
 \hline
~\cite{lopez2019deep} & RGB Image & $\checkmark$ & $\checkmark$ & $\checkmark$ & $\checkmark$ & $\times$ & $\times$ & $\times$ & $\times$ & $\times$ & $\times$ \\
\hline
~\cite{scholler2019targetless} & RGB Image, Projected Radar Data & $\checkmark$ & $\checkmark$ & $\times$ & $\times$ & $\checkmark$ & $\times$ & $\times$ & $\times$ & $\times$ & $\times$\\
\hline
~\cite{iyer2018calibnet} & RGB Image, Raw LiDAR point cloud & $\times$ & $\times$ & $\times$ & $\times$ & $\times$ & $\checkmark$ & $\checkmark$ & $\times$ & $\times$ & $\times$\\
\hline
~\cite{poursaeed2018deep} & Stereo Image pair & $\times$ & $\times$ & $\times$ & $\times$ & $\times$ & $\times$ & $\times$ & $\checkmark$ & $\times$ & $\times$\\
\hline
~\cite{en2018rpnet, melekhov2017relative, charco2018deep, hansen2012online}  & RGB Image pair & $\times$ & $\times$ & $\times$ & $\times$ & $\times$ & $\checkmark$ & $\checkmark$ & $\times$ & $\times$ & $\times$ \\
\hline
~\cite{bogdan2018deepcalib} & RGB Image & $\times$ & $\times$ & $\checkmark$ & $\checkmark$ & $\times$ & $\times$ & $\times$ & $\times$ & $\times$ & $\times$ \\
\hline
~\cite{bhardwaj2018autocalib, xiang2017posecnn, nakajima2017robust, komorowski2012extrinsic, kendall2015posenet} & RGB Image & $\times$ & $\times$ & $\times$ & $\times$ & $\times$ & $\checkmark$ & $\checkmark$ & $\times$ & $\times$ & $\times$ \\
\hline
~\cite{hold2018perceptual}  & RGB Image & \checkmark & \checkmark & \checkmark &  $\times$ & $\times$ & $\times$ & $\times$ & $\times$ & $\times$ & $\times$ \\
\hline
~\cite{workman2015deepfocal} & RGB Image & $\times$ & $\times$ & \checkmark & $\times$ & $\times$ & $\times$ & $\times$ & $\times$ & $\times$ & $\times$ \\
\hline
~\cite{shalnov2017convolutional} & Head Detections, Focal Length & $\times$ & $\times$ & $\times$ & $\times$ & $\times$ & \checkmark & \checkmark & $\times$ & $\times$ & $\times$ \\
\hline
~\cite{lee2012camera} & RGB Image & $\times$ & $\times$ & \checkmark & $\times$ & $\times$ & $\times$ & $\times$ & $\times$ & \checkmark & \checkmark \\
\hline
~\cite{ranftl2018deep} & Putative matches & $\times$ & $\times$ & $\times$ & $\times$ & $\times$ & $\times$ & $\times$ & \checkmark & $\times$ & $\times$ \\
\hline
\end{tabular}
\end{adjustbox}
\label{table:occ}
}
\end{table*}

When the camera model can be fit by enough geometric constraints, relatively exact calibration of cameras is possible thanks to the well-understood nature of picture generation and extensive research in computer vision ~\cite{hartley2003multiple}. The best results come from photographs that show man-made surroundings since they provide strong indicators like straight lines and vanishing points that can be used to determine the camera's settings. Images taken outside of a lab can also be processed using geometrical approaches. ~\cite{caprile1990using,deutscher2002automatic}. Geometric-based techniques, on the other hand, are less resistant to photographs shot in unstructured locations, with subpar equipment, or under challenging lighting circumstances since they rely on recognising and processing particular indications like straight lines and vanishing points.

A perception system needs a variety of on-board sensors to function well. Robots are using an increasing number of sensors in many modalities. A 3D LiDAR, for instance, is used in conjunction with 2D cameras in an autonomous vehicle since the latter's rich colour information complements the former's sparse distance information. Several calibration methods have been put forth recently~\cite{unnikrishnan2005fast,geiger2012automatic,levinson2013automatic,pandey2012automatic,taylor2015motion}. However, the majority of these techniques use checkerboards as fixed calibration targets, and demand a substantial amount of manual labour~\cite{unnikrishnan2005fast,geiger2012automatic}. Therefore, automatic calibration approaches that can considerably increase the flexibility and adaptability of these systems are very necessary. There have been some published methods in this field in the past~\cite{levinson2013automatic,pandey2012automatic,taylor2015motion} although the majority of these methods still rely on precise calibration parameter initialization~\cite{levinson2013automatic,pandey2012automatic}, or need a lot more information on ego-motion~\cite{taylor2015motion}.

 The majority of currently used techniques typically don't take into account the camera model's underlying mathematical formulation, instead proposing an all-encompassing framework to calculate the required parameters directly~\cite{workman2015deepfocal,rong2016radial,hold2018perceptual, lopez2019deep, zhai2016detecting, detone2016deep,bogdan2018deepcalib,zhang2020deepptz,workman2016horizon,barreto2006unifying}. Since they can only utilizing a single image, determine the camera's focal length, they are difficult to understand for practical purposes.~\cite{detone2016deep,bogdan2018deepcalib,zhang2020deepptz}. 
 The following are the main benefits of our efforts:

\begin{itemize}
    \item To the best of our knowledge, this research represents the first learning-based approach to jointly estimate both intrinsic and extrinsic camera parameters, such as camera baseline, disparity, pitch, translation, focal length, and principal point offset.
    \item The available learning-based techniques~\cite{detone2016deep, workman2015deepfocal, zhang2020deepptz} due to a paucity of data, have not been used to estimate all of the $10$ camera parameters. To resolve this difficulty, we created a synthetic dataset using data from two CARLA towns~\cite{Dosovitskiy17} featuring 49 distinct camera setups.
    \item Instead of developing a whole system to anticipate the needed parameters, unlike existing approaches~\cite{detone2016deep}, In a multi-task learning (MTL) framework, we developed a new illustration that depicts the equations for the camera model as a neural network.
    \item We put out a brand-new camera projection loss (CPL) that integrates mathematical equations into a framework for learning. 
\end{itemize}

\begin{figure}
\centering
    {\includegraphics[trim={5cm 1cm 16cm 0},clip,width=0.35\textwidth]{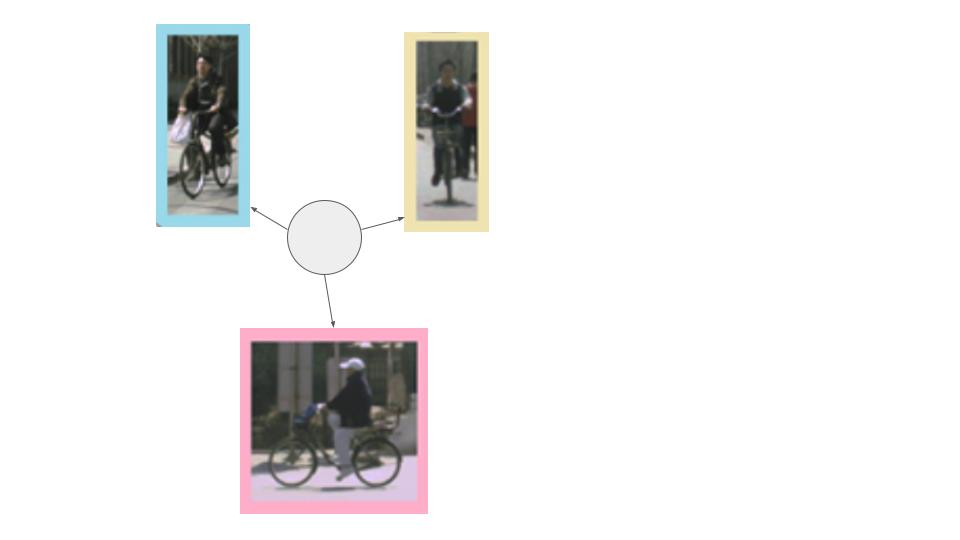}} 
    \caption{Cyclists are divided into three classes: wide (red), intermediate
(blue) and narrow (brown).~\cite{li2016new}}
    \label{fig:cdbc}
\end{figure}

\section{Related Work}
\label{sec:rw}
Table~\ref{table:occ} provides a summary of several recent settings for various camera calibration aspects. Recent studies have suggested employing learning techniques to estimate camera parameters, building on the success of convolutional neural networks. In the past, various aspects of the calibration utilising photographs problem have been researched. Workman \emph{et al.}~\cite{workman2015deepfocal} trained a CNN to do regression of a pinhole camera's field of view, concentrating on horizon line detection afterwards ~\cite{workman2016horizon}, which, provided the focal length is known, serves as a substitute for the camera's tilt and roll angles. Rong~\emph{et al.}~\cite{rong2016radial} calibrate the single-parameter radial distortion model using a classification method from Fitzgibbon~\cite{fitzgibbon2001simultaneous}. Hold-Geoffroy~\emph{et al.}~\cite{hold2018perceptual} extrinsic and intrinsic calibration were first coupled in a single network, using a classification method to predict the pinhole camera's tilt, roll, and focal length. They utilised upright 360-degree imagery, a technique that has already been used to provide training data, to artificially create any size, focal length, and rotation of the images~\cite{lopez2019deep}. Methods based on learning and tradition can coexist. In~\cite{zhai2016detecting}, The execution time and robustness compared to entirely geometric approaches are accelerated by employing learning methods to get a previous distribution on prospective camera settings is used to refine them using conventional techniques. In our paper, we do not adopt such a methodology. However, in these pipelines, the prediction made by our technique can be used as a prior. More recently, Deep-Calib~\cite{bogdan2018deepcalib,lopez2019deep} by creating distorted images with the unified projection model, incorporate the single picture self-calibration problem's distortion parameter. ~\cite{barreto2006unifying}.

In Machine Learning (ML), we often train a single model or a collection of models to solve the problem in this way. Whether it's a score based on a benchmark or a key performance indicator (KPI) for a firm, we care about optimising for a specific measure. After that, the models are adjusted and modified until their performance reaches a plateau.
This method usually yields good results, but we miss out on information that could have helped us do better on the metric that matters to us. These data were created using training signals from related activity. By swapping representations between tasks that are comparable to our starting work, we can increase the generalisation of our model.
Hard parameter sharing is the MTL strategy that neural networks use the most frequently. In most cases, this is done by retaining a small number of output layers that are task-specific while sharing the hidden layers across all jobs.
When hard parameters are shared, overfitting is significantly decreased. This makes sense logically: the more tasks we learn concurrently, the more work our model needs to undertake to build a representation that covers all of them, and the lower the likelihood we have of overfitting our starting task.
On the other hand, soft parameter sharing means that each work has its own model and unique set of parameters. Then, to promote parameter similarity, the difference between the model's parameters is regularised. Hard parameter sharing has been used in this work for learning.

Particularly, estimate of intrinsic or extrinsic parameters has been the focus of most previous applications of calibration employing multiple inputs except ~\cite{poursaeed2018deep}, when a pair of stereo pictures are used to directly estimate the fundamental matrix.  ~\cite{scholler2019targetless} to calculate rotation angles, combine radar projection data with an RGB image, ~\cite{melekhov2017relative} focus on estimating the relative camera pose from a pair of rgb photos. ~\cite{iyer2018calibnet} The 2D camera and 3D LiDAR's extrinsic calibration parameters are estimated using a rgb image from a raw LiDAR point cloud and a calibrated camera as input. ~\cite{charco2018deep} estimate the extrinsic parameters of two cameras by feeding their model two input photos that capture a scene from various viewpoints.  ~\cite{en2018rpnet} without the use of camera intrinsic or external information, takes two photos as input and immediately infers the relative positions. ~\cite{zhang2011camera} works in a variety of real-world situations, including calibrating camera intrinsic characteristics and (potentially considerable) lens distortion employing numerous photos of a known pattern, countless photographs of an unknown pattern, numerous photographs of a variety of patterns, etc.

The loss function used to calibrate a convolutional neural network is a composite loss that includes a loss component for each parameter. Typically, this situation is referred to as multi-task learning~\cite{caruana1997multitask}. The difficulty of preparing a network to do a variety of jobs with varying losses is a problem that multi-task learning must solve. The majority of these strategies use a weighted average of the loss factors; however, they vary in how the weights are determined during training: Kendall~\emph{et al.}~\cite{kendall2018multi} To weight the various loss components in accordance with a task-dependent uncertainty, use softmax and gaussian likelihoods (for regression and classification, respectively). Unlike these strategies based on uncertainty, Chen~\emph{et al.}~\cite{chen2017gradnorm}
by varying the magnitudes of the gradients connected to each loss term, one can estimate the weights' value.

\subsection{Handcrafted Features Based Methods}

~\cite{lee2012camera} outlines a way to determine camera parameters such as focal length, position, and orientation from a single photograph of a scene rectangle with an unknown aspect ratio and size. They start by resolving the unique situation in which a scene rectangle's centre is projected onto the image's centre. Coupled line cameras are used to formulate this problem, and they then give an analytical solution. They then resolve the general situation without the centering requirement by adding a quick preprocessing step. They also offer a way to determine whether a quadrilateral image is a projection of a rectangle. They use simulated and actual data to show how well the suggested strategy works.~\cite{hansen2012online} describe an approach for an online, continuous, per-frame stereo extrinsic re-calibration that can only be used with sparse stereo correspondences. They are able to model time-dependent variations because they are able to retrieve the fixed baseline's five degrees of freedom extrinsic position for each frame. Minimizing epipolar errors yields the initial extrinsic estimates, which are then filtered with a Kalman filter (KF). The lower bound of the solution uncertainty calculated by Crámer-Rao is used to calculate observation covariances.~\cite{komorowski2012extrinsic} explains how to estimate the rotation and translation matrices of a camera from a series of photos. It is presumptively known and fixed camera intrinsic matrix and distortion coefficients throughout the entire sequence.~\cite{geiger2012automatic} a web-based toolset for fully automatic calibration of cameras and distances between cameras. Since most calibrations just require a single image and range scan cases,in the event of ambiguities, the suggested camera-to-range registration strategy is capable of finding a variety of solutions., making it robust to a variety of simple to use, fully automatic, and imaging conditions. The durability of their system is demonstrated in tests utilising a number of sensors, including as colour and grayscale cameras, the Kinect 3D sensor, and the Velodyne HDL-64 laser scanner, in both indoor and outdoor settings, and in various lighting situations.~\cite{zhang2011camera} describe a technique for calibrating a camera's inherent properties and (potentially severe) lens distortion. Furthermore, this approach does not depend on low-level features like edges or corners for feature extraction. The approach makes use of cutting-edge tools for high-dimensional convex optimization, especially those for recovering sparse signals and matrix rank minimization. They demonstrate how Principal Component Analysis might be extended to address the camera calibration problem and resolved using analogous methods. ~\cite{strecha2008benchmarking} initiate a conversation about whether image-based 3-D modelling methods could take the place of LIDAR equipment in the collection of outside 3D data. In this context, the internal and exterior camera calibration, as well as dense multi-view stereo, are the two key challenges that need to be addressed. They have collected test results from outdoor LIDAR and camera situations to evaluate both.~\cite{furukawa2009accurate} presents a method for camera calibration that effectively directs utilising a common bundle adjustment approach, the search for further image correspondences and greatly improves camera calibration parameters. This method uses imprecise camera parameter estimates and the results of a multi-view stereo system on resized input photos provide top-down information.~\cite{yokochi2006extrinsic} proposes a unique technique for predicting extrinsic camera parameters using sparse GPS position data and feature points from an image sequence. Their approach, which is enhanced by the use of GPS data to minimise accumulative estimating errors and is based on a structure-from-motion technique. Additionally, mistracked features are also removed using the position data. The suggested technique enables estimation of extrinsic parameters from extremely lengthy image sequences without accumulation of errors.~\cite{xu2002determining} determines both the intrinsic and extrinsic camera settings while simultaneously using numerous photos of multiple balls. Conics or ellipses that resemble balls are projected onto the image. There are 5 separate constraints for each conic. 30 limitations are provided by 3 balls in 2 photos. There are 12 parameters for the placements and radii of the three balls, and there are 15 intrinsic and extrinsic parameters altogether. Therefore, the bare minimum configuration to tackle this problem is 3 balls in 2 photos. The ball placements and radii, as well as any unknown camera parameters, are best inferred from the boundary points. The algorithm's effectiveness and reliability are demonstrated by experimental results on real photos. Calibrating network in ~\cite{ahmed1999neurocalibration} can identify the matrix of perspective, projection, and transformation between relevant 2D picture pixels and the 3D world points. The parameters of a camera model can be specified online that fulfil the orthogonality restrictions on the rotational transformation starting with random initial weights.

\begin{figure}
\centering
    {\includegraphics[trim={0 0 0 0},clip,width=1.0\columnwidth]{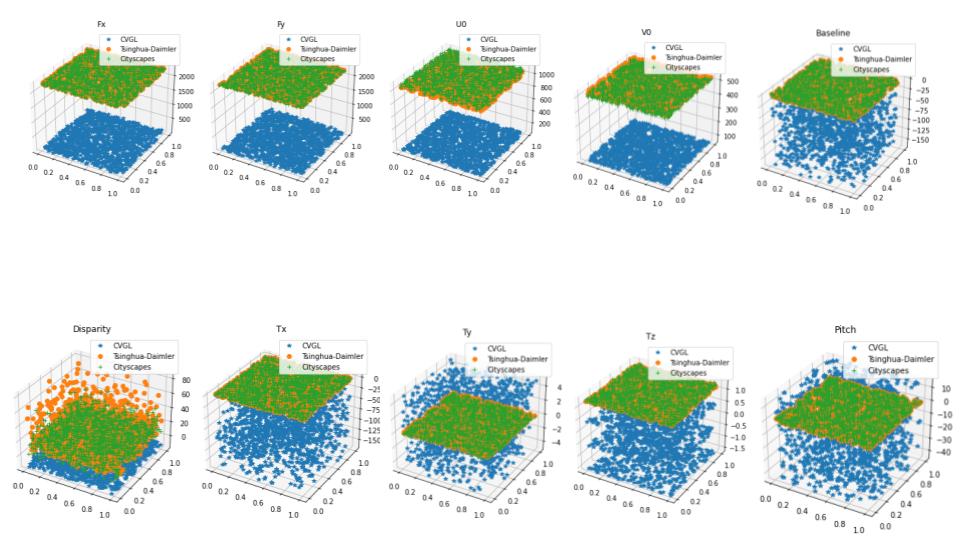}} 
    \caption{Statistical Comparison of Camera Parameters from 3 datasets used. Blue colour denotes CVGL Camera Calibration Dataset while Green colour represents Cityscapes Dataset and Orange colour has been used for Tsinghua-Daimler Cyclist Detection Benchmark. The plots follow the following order from left to right: $f_x, f_y, u_0, v_0, b, d, t_x, t_y, t_z, \theta_p$.}
    \label{fig:stats}
\end{figure}

\begin{table*}[t]

\caption{Table showing statistics of the used datasets.}
\centering
{
\begin{adjustbox}{max width=\textwidth}
\begin{tabular}{ |l|c|c|c|c|c|c|c|c|c|c|c| } 
\hline
Dataset &Value& $f_x$ & $f_y$ & $u_0$ & $v_0$ & $b$ & $d$ & $t_x$ & $t_y$ & $t_z$ & $\theta_p$\\
 \hline
\multirow{3}{4em}{CVGL}&Min & 15.005 & 15.005 & 56.0 & 56.0 & -168.0 & -16.0 & -168.0 & -5.0 & -1.6 & -45.0\\
&Mean & 59.209 & 59.209 & 56.0 & 56.0 & -83.870 & -10.966 & -83.870 & 0.425 & -0.572 & -12.235\\
&Max & 120.092 & 120.092 & 56.0 & 56.0 & 0.0 & 14.531 & 0.0 & 5.0 & 0.4 & 15.0\\
\hline
\multirow{3}{4em}{Tsinghua-Daimler}&Min & 2282.864 & 2281.794 & 1042.041 & 529.888 & 0.208 & -6.753 & 2.0 & 0.125 & 1.23 & 0.022\\
&Mean & 2282.864 & 2281.794 & 1042.041 & 529.888 & 0.208 & 21.455 & 2.0 & 0.125 & 1.23 & 0.021\\
&Max & 2282.864 & 2281.794 & 1042.041 & 529.888 & 0.208 & 83.093 & 2.0 & 0.125 & 1.23 & 0.022\\
\hline
\multirow{3}{4em}{Cityscapes}&Min & 2262.52 & 2225.540 & 1045.53 & 513.137 & 0.209 & -4.675 & 1.7 & -0.1 & 1.18 & 0.038\\
&Mean & 2264.256 & 2253.738 & 1081.188 & 514.988 & 0.213 & 15.539 & 1.699 & 0.037 & 1.230 & 0.041\\
&Max & 2268.36 & 2265.301 & 1096.98 & 519.277 & 0.222 & 57.339 & 1.7 & 0.1 & 1.3 & 0.05\\
\hline
\end{tabular}
\end{adjustbox}
\label{table:dataset_stats}
}
\end{table*}

\subsection{CNN Based Methods}

~\cite{lopez2019deep} describe a technique for extrapolating from a single image both inherent (radial distortion and focal length) and extrinsic (tilt and roll) properties. Instead of simply predicting the distortion parameters, they suggest a more learner-friendly parameterization for radial distortion. Additionally, predicting more heterogeneous factors makes the difficulty of loss balancing worse. Preventing the need to reconcile heterogeneous loss terms, they suggest a novel point projection-based loss function. They are the first to use a single image to simultaneously determine the focal length, radial distortion, tilt, roll, and other parameters.~\cite{scholler2019targetless} put forth the first data-driven technique for automatic rotating radar-camera calibration without specific calibration objectives. A coarse and a fine convolutional neural network form the foundation of their strategy. They use a training approach that takes inspiration from boosting and trains the fine network using the residual error of the coarse network. They gathered their own real-world data because there were no public databases combining radar and camera observations.~\cite{iyer2018calibnet}, in order to automatically estimate the 6-DoF rigid body switch in real time between a 2D camera and a 3D LiDAR, they propose a geometrically supervised deep network. CalibNet eliminates the requirement for calibration targets, which significantly reduces the amount of calibration work required. The network only accepts the corresponding monocular image, the camera calibration matrix K, and a LiDAR point cloud as inputs during training. They don't enforce direct control throughout train time (i.e., For instance, they do not directly regress to the calibration parameters. Instead, they train the network to predict calibration parameters that, from a geometric and photometric perspective, maximise the consistency of the input point clouds and images. Without retraining or domain adaption, CalibNet gains the ability to solve the fundamental geometry problem iteratively while properly predicting criteria for extrinsic calibration for various miscalibrations.~\cite{poursaeed2018deep}, in order to estimate fundamental matrices they propose new neural network topologies that can be extended end-to-end without relying on point correspondences. To maintain the fundamental matrix's mathematical characteristics as a seven-degree-of-freedom homogeneous rank-2 matrix, new modules and layers are added.~\cite{en2018rpnet}, without relying on camera intrinsic or external information, the proposed RPNet network receives pairs of photos as input and infers the relative positions directly.~\cite{charco2018deep} suggests using a deep learning network architecture for multi-view environment relative camera position estimation. The suggested network uses an AlexNet variant architecture as a regressor to anticipate the output of relative translation and rotation. The suggested method is trained from scratch on a sizable data set using two photos from the same scene as input.~\cite{ranftl2018deep} creates a model for the issue that divides it into a number of weighted homogeneous least-squares issues, with robust weights calculated by deep networks. The proposed approach performs feature extraction, matching, and model fitting directly on putative correspondences and so integrates into common 3D vision workflows. The method produces robust estimators that are computationally effective and can be trained from beginning to end.~\cite{bogdan2018deepcalib} automatically calculates, from a single input image, the focal length and distortion parameter of the camera. In order to automatically create a massive dataset of millions of wide field-of-view photos with ground truth intrinsic parameters, they use the abundance of omnidirectional images that are readily available on the Internet to train the CNN.~\cite{bhardwaj2018autocalib} a method for scalable, automatic calibration of traffic cameras called AutoCalib. In order to automatically produce a reliable estimate of the camera calibration parameters from just hundreds of samples, AutoCalib utilises deep learning to extract specific key-point features from video clips of cars. The camera calibration parameters are output when a video segment has been input.~\cite{xiang2017posecnn} uses a single RGB picture to predict translation and rotation.~\cite{melekhov2017relative} proposes a method for calculating the relative posture between two cameras that is based on convolutional neural networks. The proposed network immediately generates the relative rotation and translation as output after receiving RGB images from both cameras as input. Transfer learning is used to train the system from beginning to end using a sizable classification dataset. The findings show that the proposed method clearly outperforms the baseline when compared to popular local feature-based methods (SURF, ORB). A variation of the suggested architecture with a spatial pyramid pooling (SPP) layer is also tested and found to enhance performance even more.~\cite{hold2018perceptual}, a deep convolutional neural network can directly extract camera calibration parameters from a single image. This network outperforms previous techniques, including new deep learning-based systems, in terms of standard L2 error and is trained using autonomously generated samples from a sizable panorama dataset. Users are asked to assess the realism of 3D objects composited with and without ground truth camera calibration as part of a sizable human perception study that is being conducted. They build a new perceptual measure for camera calibration on the basis of this research, and they show that their deep calibration network performs better on this measure than existing approaches. Finally, they show how to use their calibration network for a variety of tasks, such as picture retrieval, compositing, and the insertion of virtual objects.~\cite{shalnov2017convolutional} based on seeing people in the input video that was shot with a static camera, provide a novel approach to estimating camera posture. In particular, the suggested solution treats the used object detector as a parameter rather than making any assumptions about it. Additionally, they just train on synthetic data and do not require a sizable labelled collection of actual data.~\cite{nakajima2017robust}, using several feature descriptor databases created for each partitioned perspective and a feature descriptor that is almost invariant for each keypoint, we present a unique approach to robust camera pose estimation. Based on a set of training photos created for each perspective class, their method uses deep learning to predict the viewpoint class for each incoming image. For preparing these photos for deep learning and creating databases, they provide two different methods. Images are created using a projection matrix in the first technique to ensure reliable learning in a variety of contexts with shifting backdrops. The second technique makes use of actual visuals to teach a planar pattern in a specific setting.~\cite{workman2015deepfocal}, to directly estimate the focal length using just raw pixel intensities as input features, investigate the application of a deep convolutional neural network trained on natural photos acquired from online photo libraries.~\cite{kendall2015posenet}, rotation and translation predictions are made using an RGB picture.~\cite{zhang2020deepptz} suggest a deep learning-based method for automatically estimating the rotation angles, distortion parameters, and focal lengths.

\begin{figure}
\centering
    {\includegraphics[trim={0 12cm 0 0},clip,width=1.0\columnwidth]{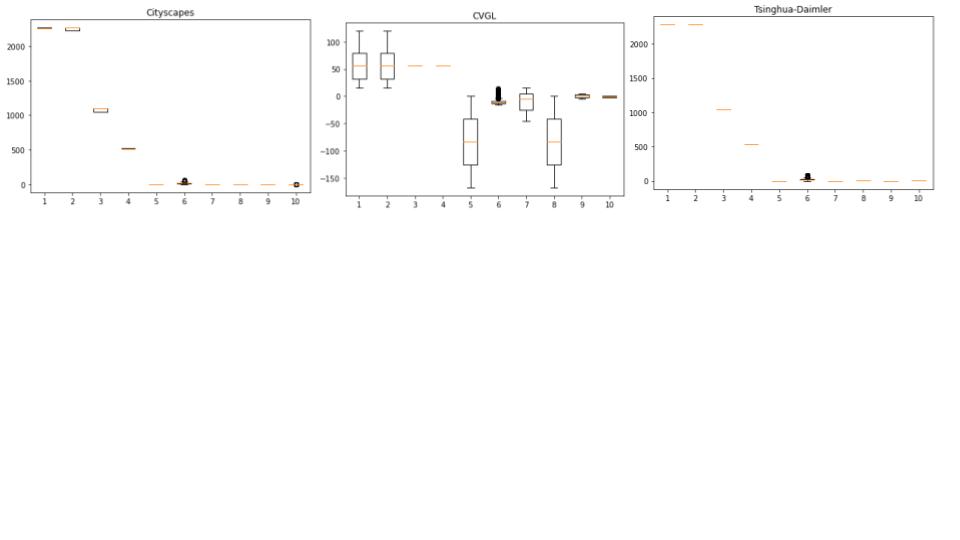}} 
    \caption{Box plots of all the parameters in Cityscapes, CVGL and Tsinghua-Daimler dataset respectively. The plots follow the following order from left to right: $f_x, f_y, u_0, v_0, b, d, \theta_p, t_x, t_y, t_z$.}
    \label{fig:plots}
\end{figure}

\subsection{Deep-unfolding Based Methods}

Deep-unfolding caters to incorporating mathematics in the architecture. ~\cite{poursaeed2018deep}, to maintain the underlying matrix's mathematical characteristics as a seven-degree-of-freedom homogeneous rank-2 matrix, suggest new modules and layers. They add a reconstruction layer and a normalising layer. The reconstruction layer produces a fundamental matrix after receiving parameters as input. With this method, rank two of the reconstructed matrix is assured. The F-matrix is typically normalised by dividing it by the last entry.~\cite{nie2021depth}, to estimate homography, introduce a contextual correlation layer (CCL). The CCL can be employed in a variety of ways in a learning framework and is effective at capturing the long-range correlation found in feature maps.~\cite{ye2021motion} suggest using the weighted average, one may create a homography flow representation. of the eight previously established homography flow bases. They suggest a Low Rank Representation (LRR) block that reduces the feature rank so that characteristics associated with the dominating motions are kept and other features are rejected. They take into account that 8 Degrees of Freedom (DOFs) are present in a homography, which is much fewer than the rank of the network features. They suggest that by performing a feature identity loss (FIL) to verify that the learned image features are warp-equivariant—that is, that the results should remain the same when the warp operation and feature extraction are carried out in the opposite order—we may prevent this from happening.

When feasible, task-agnostic techniques can be replaced with domain knowledge to balance loss components: Yin et al.~\cite{yin2018fisheyerecnet}, calibration of an 8-parameter fisheye lens distortion model should be done. When attempting to minimise the parameter directly, they point out the challenge of balancing loss components of various natures and suggest a different approach based on the in accordance with the photometric inaccuracy. In this study, we additionally investigate the problem of balancing the loss components for camera calibration and propose a speedier approach based on camera projections, similar to DeepCalib~\cite{lopez2019deep} nonetheless, employing dependent proxy variables through a 2D to 3D projection as opposed to using proxy variables (that can be seen in the image directly) such as horizon line and the distance between the image's horizon line and its centre. An overview of our proposed approach is presented in Fig.~\ref{fig:flowdiagram}

\begin{figure}
\centering
    {\includegraphics[trim={0 12cm 23cm 4cm},clip,width=0.75\columnwidth]{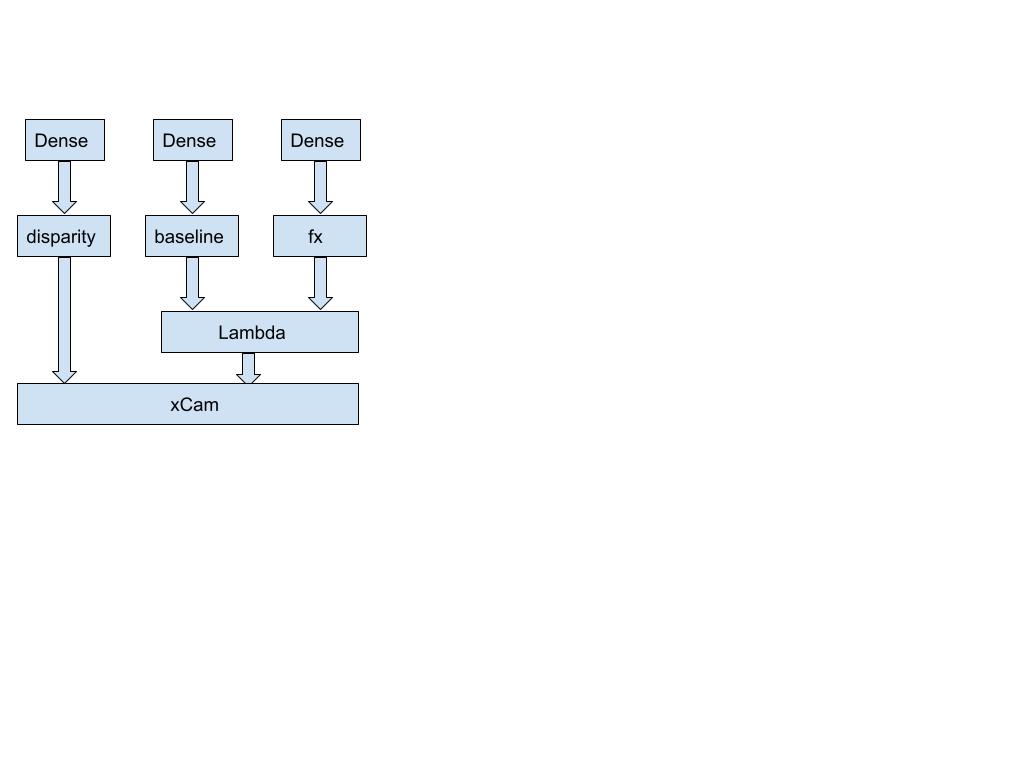}} 
    \caption{Lambda layer representation of $x_{cam}$.}
    \label{fig:xcam}
\end{figure}

\begin{figure}
\centering
    {\includegraphics[trim={0 4cm 10cm 4cm},clip,width=0.75\columnwidth]{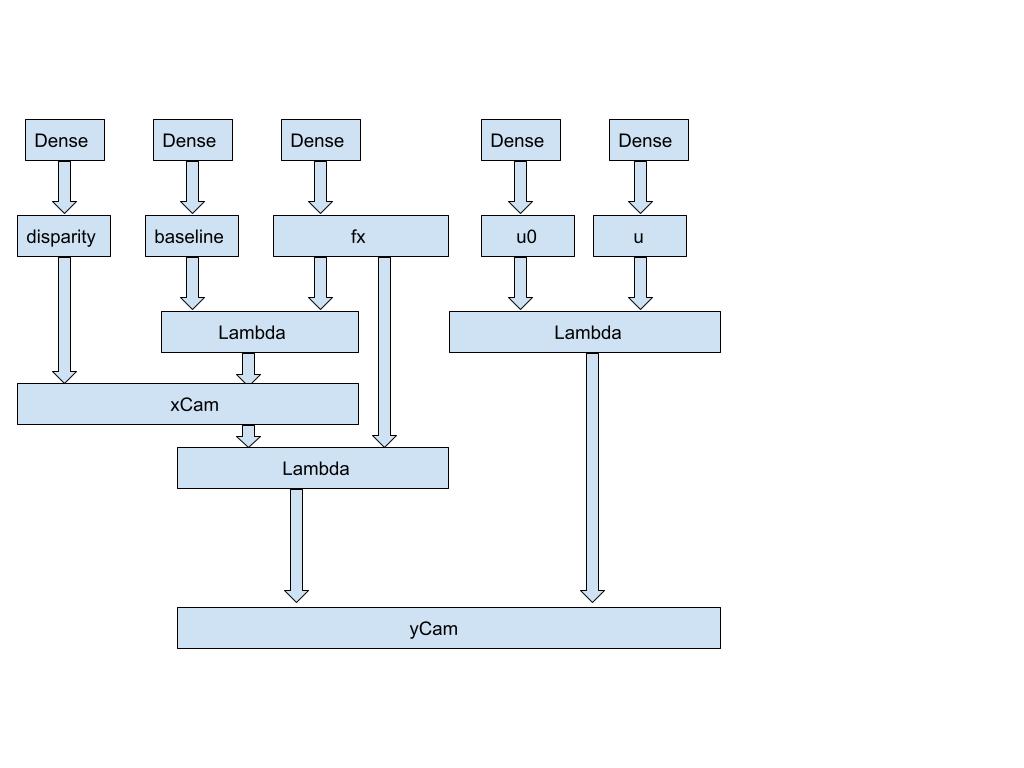}} 
    \caption{Lambda layer representation of $y_{cam}$.}
    \label{fig:ycam}
\end{figure}

\begin{figure}
\centering
    {\includegraphics[trim={0 10cm 7cm 4cm},clip,width=0.75\columnwidth]{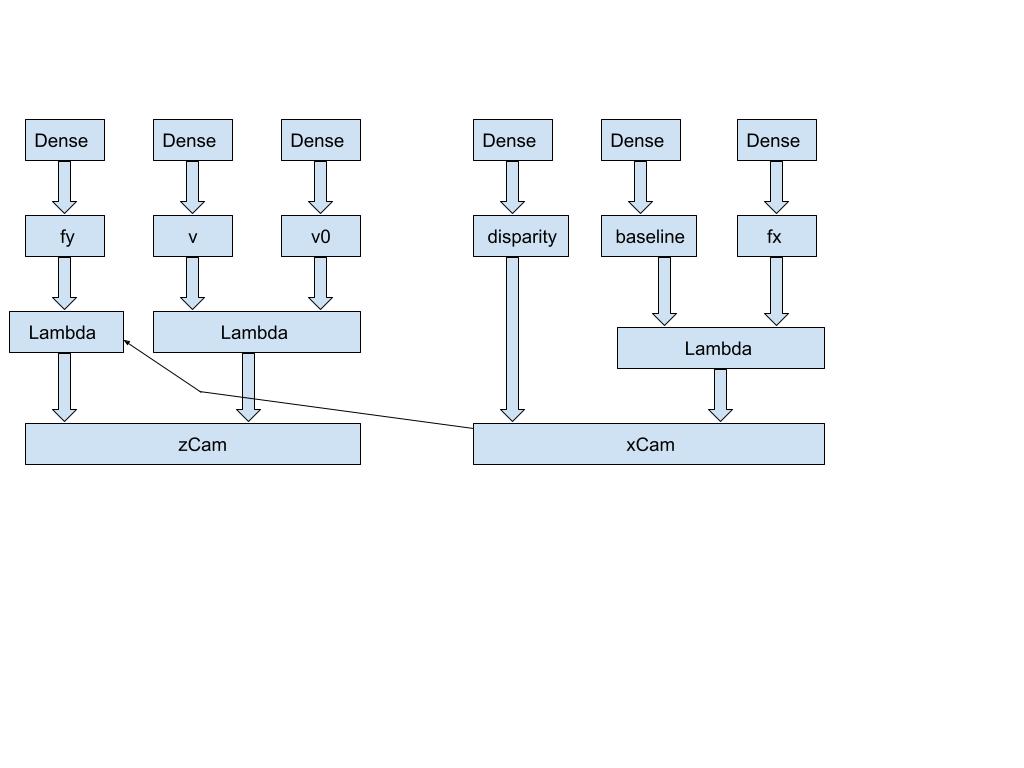}} 
    \caption{Lambda layer representation of $z_{cam}$.}
    \label{fig:zcam}
\end{figure}

\begin{figure}
\centering
    {\includegraphics[trim={0 10cm 13cm 3cm},clip,width=1.0\columnwidth]{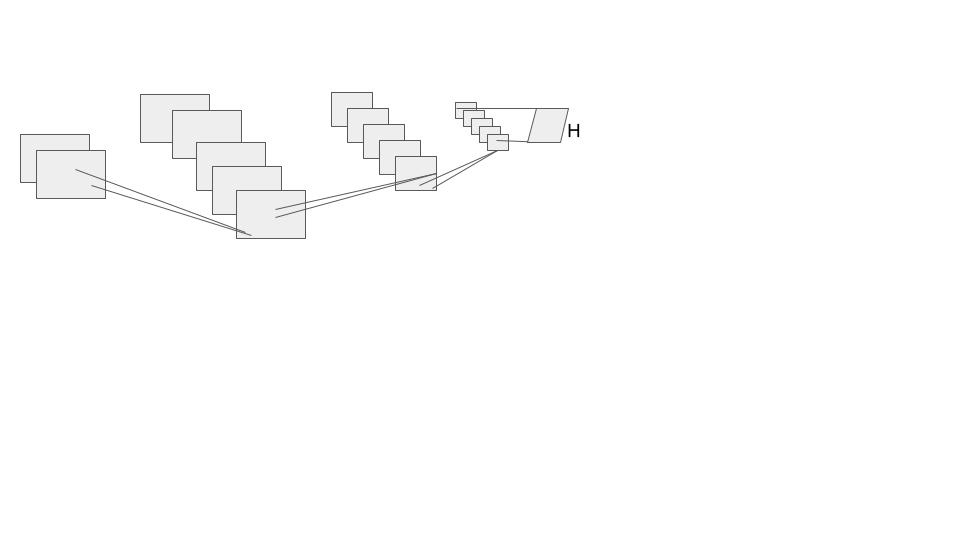}} 
    \caption{Deep Image Homography Estimation. HomographyNet, a Deep Convolutional Neural Network, calculates the homography between two images directly. Corner identification and homography estimation are not required as distinct phases in the method, and all parameters are trained end-to-end using a sizable dataset of labelled pictures.~\cite{detone2016deep}}
    \label{fig:dh}
\end{figure}

\section{Proposed Method}
\label{sec:pm}


In order to improve representation while learning, we suggest using multi-task learning using an unique loss function built into a neural network to add mathematical equations.
In order to forecast the intrinsic and extrinsic camera characteristics, a convolutional neural network is trained. As the feature extractor, we employ dependent regressors with a comparable network design to accomplish this. We use a Inception-v3~\cite{szegedy2016rethinking} pretrained on ImageNet~\cite{russakovsky2015imagenet}
as a feature extractor, the Lambda layers for loss computation are then applied, using 13 regressors, of which 3 are related to the 3D point cloud and 3 are related to the camera settings. We use proxy variables, which are dependent on one another but are not visible in the image, to forecast the translation, baseline, pitch, primary point, focal length, instead of training these regressors to do so. This enables us to connect our approach directly to the multi-view geometry's mathematical underpinnings~\cite{Hartley:2003:MVG:861369} resulting in better performance. 

\textbf{Camera Model}
The perspective projection model based on the pinhole camera is the camera model that we take into consideration~\cite{faugeras1993three}.
If $M$ has world coordinates $(X, Y, Z)$ and projects onto a point $m$ that has pixel coordinates $(u, v)$, the operation can be described, in homogeneous coordinates, by the equation:
\begin{align}
 S\begin{pmatrix} u \\ v \\ 1 \end{pmatrix} = \mathbf{P}\begin{pmatrix} X \\ Y \\ Z \\ 1 \end{pmatrix}
 \label{eq:1}
  \end{align}
where $S$ is a scaling factor and the matrix $\mathbf{P}$ is in the
format
\begin{align}
 P = \begin{pmatrix} p_1^T & p_{14} \\ p_2^T & p_{24}\\ p_3^T & p_{34} \end{pmatrix}
 \label{eq:2}
  \end{align}
The $3 \times 4$ matrix $\mathbf{P}$ is commonly referred to as perspective projection matrix and decomposed into two
matrices: $\mathbf{P} = \mathbf{A}\mathbf{D}$ where
\begin{align*}
 \mathbf{D} = \begin{pmatrix} \mathbf{R} & t \\ 0_3^T & 1\end{pmatrix} \
 \mathbf{A} = \begin{pmatrix} \alpha_u & -\alpha_u\cot\theta  & u_0 & 0\\ 0 & \frac{\alpha_v}{\sin\theta} & v_0 & 0\\0 & 0 & 1 & 0\end{pmatrix}
\end{align*}
The $4 \times 4$ matrix $D$ represents the mapping from
world coordinates to camera coordinates and accounts
for six extrinsic parameters of the camera: three for
the rotation $\mathbf{R}$ which is normally specified by three
rotation (Euler) angles: $ R_x,\:R_y,\:R_z $ and three for
the translation $ t=(t_x,\:t_y,\:t_z)^T.$ $0_3$ represents the null vector $(0,\:0,\:0)^T.$ The $3 \times 4$ matrix $\mathbf{A}$ represents the
intrinsic parameters of the camera: the scale factors $\alpha_u \:{and} \:\alpha_v$, the coordinates $u_0 \:{and}\: v_0$ of the principal
point, and the angle $\theta$ between the image axes.

In this work, we use 2D to 3D projection as a frame of reference, leaving $13$ free parameters: focal length ($f_x$, $f_y$), principal point ($u_0$, $v_0$), disparity, baseline, pitch, translation ($t_x$, $t_y$, $t_z$) and 3D coordinates $(X, Y, Z)$.

Consequently, the network's required parameters are
the focal length ($f_x$, $f_y$), principal point ($u_0$, $v_0$), disparity, baseline, pitch and translation ($t_x$, $t_y$, $t_z$).

\textbf{Parameterization}
As revealed by previous work ~\cite{workman2015deepfocal,workman2016horizon,hold2018perceptual}, the convergence and final performance of the network can be considerably enhanced by a suitable parameterization of the variables to forecast. It might be challenging to deduce camera calibration characteristics from the image content, such as the focal length or tilt angles. Instead, proxy parameters that are easily visible in the image can serve as a more accurate representation of them. As a stand-in for our parameters, we project 2D data into 3D space.

$\mathbf{A}$ can also be written as:

\begin{align}
\begin{pmatrix} f_x & 0 & u_0 \\ 0 & f_y & v_0 \\ 0 & 0 & 1 \end{pmatrix}
\end{align}

A 2D point in the image coordinate system is projected to the camera coordinate system, and subsequently to the world coordinate system (Eq.~\ref{eq:1}) and (Eq.~\ref{eq:2}) as:
\begin{align}
\begin{pmatrix} u \\ v \\ 1 \end{pmatrix} \sim \begin{pmatrix} f_x & 0 & u_0 \\ 0 & f_y & v_0 \\ 0 & 0 & 1 \end{pmatrix} \begin{pmatrix} r_{11} & r_{12} & r_{13} & t_x\\ r_{21} & r_{22} & r_{23} & t_y\\ r_{31} & r_{32} & r_{33} & t_z \end{pmatrix} \begin{pmatrix} X \\ Y \\ Z \\ 1 \end{pmatrix}
 \label{eq:3}
  \end{align}

Combining (Eq.~\ref{eq:1}) , (Eq.~\ref{eq:2}) and $\mathbf{D}$ as:

\begin{align}
 \begin{pmatrix} X \\ Y \\ Z \\ 1 \end{pmatrix} \sim \begin{bmatrix}\begin{pmatrix} f_x & 0 & u_0 \\ 0 & f_y & v_0 \\ 0 & 0 & 1 \end{pmatrix} \begin{pmatrix} \mathbf{R} & t \\ 0_3^T & 1\end{pmatrix}\end{bmatrix}^{-1} \begin{pmatrix} u \\ v \\ 1 \end{pmatrix}
 \label{eq:4}
  \end{align}
\begin{align}
 \begin{pmatrix} X \\ Y \\ Z \\ 1 \end{pmatrix} \sim \begin{pmatrix} \mathbf{R} & t \\ 0_3^T & 1\end{pmatrix}^{-1} \begin{pmatrix} f_x & 0 & u_0 \\ 0 & f_y & v_0 \\ 0 & 0 & 1 \end{pmatrix}^{-1} \begin{pmatrix} u \\ v \\ 1 \end{pmatrix}
 \label{eq:5}
  \end{align}

Image to Camera Transformation can be performed as follows:

\begin{align}
\mathbf{A}^{-1} = \begin{pmatrix} \frac{1}{f_x} & 0 & \frac{-u_0}{f_x} \\ 0 & \frac{1}{f_y} & \frac{-v_0}{f_y} \\ 0 & 0 & 1 \end{pmatrix}
\end{align}

Assuming skew = 0

\begin{align}
 \begin{pmatrix} y_{cam} \\ z_{cam} \\ x_{cam} \end{pmatrix} \sim \begin{pmatrix} \frac{1}{f_x} & 0 & \frac{-u_0}{f_x} \\ 0 & \frac{1}{f_y} & \frac{-v_0}{f_y} \\ 0 & 0 & 1 \end{pmatrix} \begin{pmatrix} u \\ v \\ 1 \end{pmatrix}
 \label{eq:6}
  \end{align}
\begin{subequations}
  \begin{equation}
    \label{eq-6a}
      y_{cam} = \frac{u}{f_x} - \frac{u_0}{f_x} = \frac{u - u_0}{f_x}
  \end{equation}
  \begin{equation}
    \label{eq-6b}
    z_{cam} = \frac{v}{f_y} - \frac{v_0}{f_y} = \frac{v - v_0}{f_y}
  \end{equation}
    \begin{equation}
    \label{eq-6c}
    x_{cam} = 1
  \end{equation}
\end{subequations}

For Camera to World Transformation:

\begin{align}
 \begin{pmatrix} X \\ Y \\ Z \\ 1 \end{pmatrix} \sim \begin{pmatrix} \mathbf{R} & t \\ 0_3^T & 1\end{pmatrix} \begin{pmatrix} x_{cam} \\ y_{cam} \\ z_{cam} \\ 1 \end{pmatrix}
 \label{eq:7}
  \end{align}
\begin{align}
 \begin{pmatrix} X \\ Y \\ Z \end{pmatrix} \sim \begin{pmatrix} \cos\theta & 0 & \sin\theta \\ 0 & 1 & 0 \\ -\sin\theta & 0 & \cos\theta\end{pmatrix} \begin{pmatrix} x_{cam} \\ y_{cam} \\ z_{cam} \end{pmatrix} + \begin{pmatrix} x \\ y \\ z \end{pmatrix}
 \label{eq:8}
  \end{align}
\begin{subequations}
  \begin{equation}
    \label{eq-8a}
      X = x_{cam} * \cos\theta + z_{cam} * \sin\theta + x
  \end{equation}
  \begin{equation}
    \label{eq-8b}
    Y = y_{cam} + y
  \end{equation}
    \begin{equation}
    \label{eq-8c}
    Z = -x_{cam} * \sin\theta + z_{cam} * \cos\theta + z
  \end{equation}
\end{subequations}
%


%
%

To project a point from image to camera coordinate:
\begin{subequations}
  \begin{equation}
    \label{eq-4a}
      x_{cam} = f_x * b / disparity
  \end{equation}
  \begin{equation}
    \label{eq-4b}
    y_{cam} = -(x_{cam} / f_x) * (u - u_0)
  \end{equation}
    \begin{equation}
    \label{eq-4c}
    z_{cam} = (x_{cam} / f_y) * (v_0 - v)
  \end{equation}
\end{subequations}
$x_{cam}$ works as a proxy for $f_x$, baseline and disparity while $y_{cam}$ works as a proxy for $f_x$, u and $u_0$ and $z_{cam}$ works as a proxy for $f_y$, $v$ and $v_0$. Using $x_{cam}$, $y_{cam}$ and $z_{cam}$ from Eq.~\ref{eq-4a}, Eq.~\ref{eq-4b} and Eq.~\ref{eq-4c} respectively, points can be projected to world coordinate system using:
\begin{subequations}
  \begin{equation}
    \label{eq-5a}
      X = x_{cam} * \cos(\theta_p) + z_{cam} * \sin(\theta_p) + t_x
  \end{equation}
  \begin{equation}
    \label{eq-5b}
    Y = y_{cam} + t_y
  \end{equation}
    \begin{equation}
    \label{eq-5c}
    Z = -x_{cam} * \sin(\theta_p) + z_{cam} * \cos(\theta_p) + t_z
  \end{equation}
\end{subequations}
$X$ works as a proxy for pitch and $t_x$ while $Y$ works as a proxy for $t_y$ and $Z$ works as a proxy for pitch and $t_z$.

\begin{figure*}
\centering
    {\includegraphics[width=0.75\textwidth]{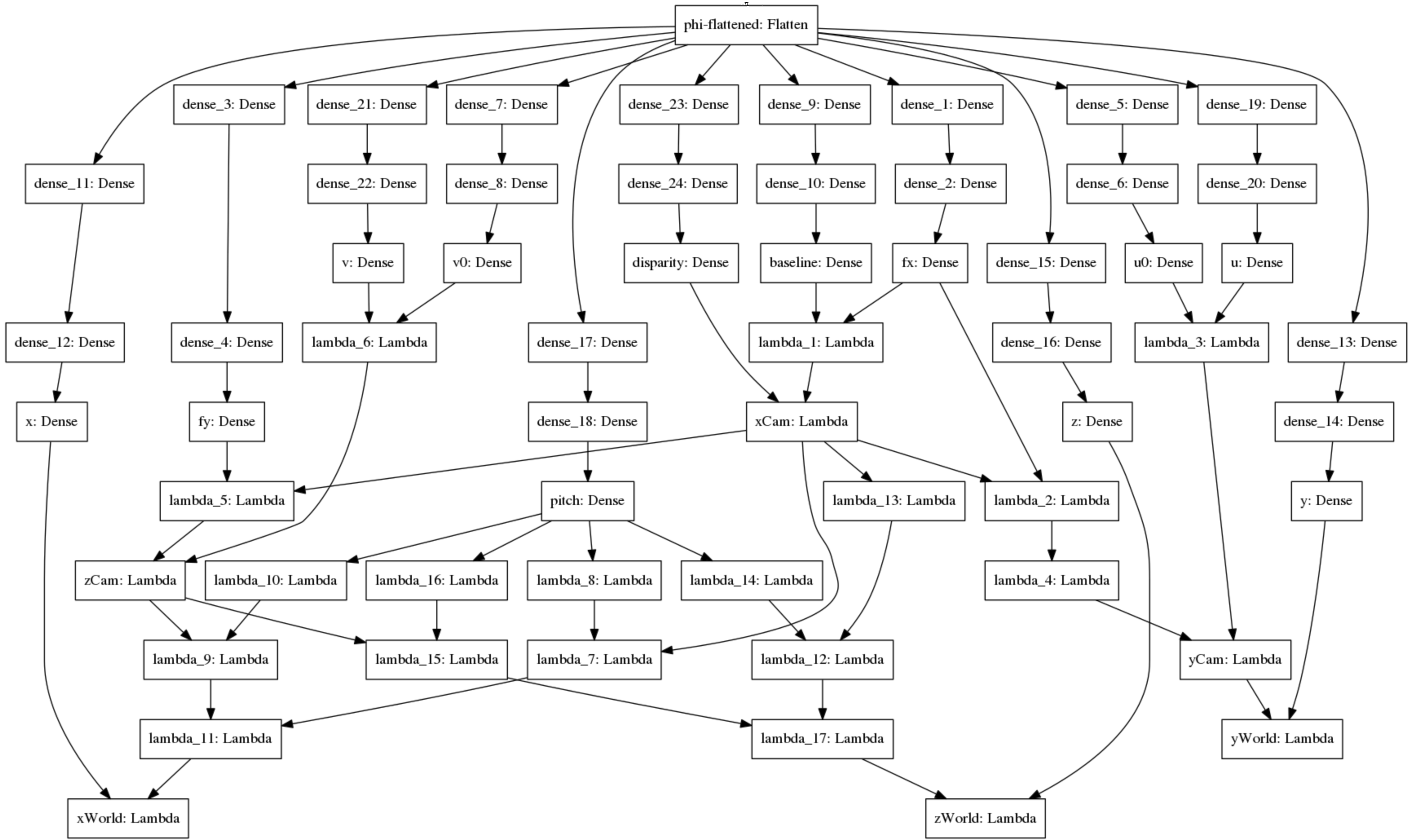}} 
    \caption{Camera Projection Loss in the form of Lambda layers. Lambda layers have been used to implement the loss using (Eq.~\ref{eq-4a} - Eq.~\ref{eq-5c}).}
    \label{fig:camera_projection_loss}
\end{figure*}

\textbf{Camera Projection Loss}: It is especially important to weigh the loss components when training a single architecture to predict parameters of various magnitudes so that the learning process is not dominated by the estimation of specific parameters.
We observe that a single metric based on the projection of points from 2D to 3D can be used for the calibration of cameras in place of individually adjusting the camera settings. Lambda layer representation of the loss can be seen in Fig.~\ref{fig:camera_projection_loss}.

Given two images with known parameters $\omega=(f_x, f_y, u_0,\\ v_0, b, d, \theta_p, t_x, t_y, t_z, X, Y, Z)$ and a prediction of such parameters
given by the network $\hat{\omega}$=($f_x^{'}, f_y^{'}, u_0^{'}, v_0^{'}$, $b^{'}$, $d^{'}$, $\theta_p^{'}$, $t_x^{'}$, $t_y^{'}$, $t_z^{'}$, $X^{'}$, $Y^{'}$, $Z^{'}$), we get the projected point in world coordinate system through Eq.~\ref{eq-4a} - Eq.~\ref{eq-5c}. Loss is computed between actual $\omega$ and predicted  $\hat{\omega}$ using: 
  \begin{equation}
    \label{eq-6}
    L({\omega}, \hat{\omega}) = (\frac{1}{n})\sum_{i=1}^{n}MAE({\omega} , \hat{\omega})
  \end{equation}

\textbf{Separating sources of loss errors}: By defining many errors in terms of a single measure, the suggested loss resolves the task balance issue. A new issue arises during learning when employing several camera characteristics to anticipate the 3D points, as a point's departure from its ideal projection can be attributed to multiple parameters. In other words, a parameter error might affect other parameters by backpropagating through the camera projection loss.

To get around this issue, we separate the camera projection loss and assess it separately for each parameter similar to~\cite{lopez2019deep}:
\begin{align*}
\begin{split}
  L_{f_x} &= L((f_x, f_y^{GT}, u_0^{GT}, v_0^{GT}, b^{GT}, d^{GT}, \theta_p^{GT},\\  &t_x^{GT}, t_y^{GT}, t_z^{GT}, X^{GT}, Y^{GT}, Z^{GT}),\omega) \\
    L_{f_y} &= L((f_x^{GT}, f_y, u_0^{GT}, v_0^{GT}, b^{GT}, d^{GT},  \theta_p^{GT},\\  &t_x^{GT}, t_y^{GT}, t_z^{GT}, X^{GT}, Y^{GT}, Z^{GT}),\omega) \\
    \ldots \\
        L_{Z} &= L((f_x^{GT}, f_y^{GT}, u_0^{GT}, v_0^{GT}, b^{GT}, d^{GT}, \theta_p^{GT},\\  &t_x^{GT}, t_y^{GT}, t_z^{GT}, X^{GT}, Y^{GT}, Z),\omega)
  \end{split}
\end{align*}
\begin{align}
\label{eq-7}
\begin{split}
  L^{*} &= \frac{L_{f_x} + L_{f_y} + L_{u_0} + ... + L_{Z}}{13}
  \end{split}
\end{align}

By adding weights, the loss function is further normalised to prevent the unneeded bias caused by one or more error terms $\alpha_i$ with each of the parameters. Due to the varied ranges of several parameters, this bias is introduced. These weights $\alpha_i$ are learned adaptively during the training process. The updated loss function is defined as:

\begin{align}
\label{eq-9}
\begin{split}
  L^{*} &= \alpha_{f_x}L_{f_x} + \alpha_{f_y}L_{f_y} + \alpha_{u_0}L_{u_0} + ... + \alpha_{Z}L_{Z}
  \end{split}
\end{align}


\begin{figure}
    \centering
    \begin{tabular}{cc}
        {\includegraphics[width=0.35\columnwidth]{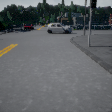}} &
        {\includegraphics[width=0.35\columnwidth]{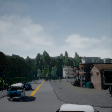}} \\
        (a) & (b) \\
        {\includegraphics[width=0.35\columnwidth]{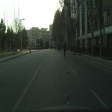}} &
        {\includegraphics[width=0.35\columnwidth]{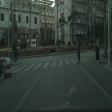}}\\
         (c) & (d)\\
        {\includegraphics[width=0.35\columnwidth]{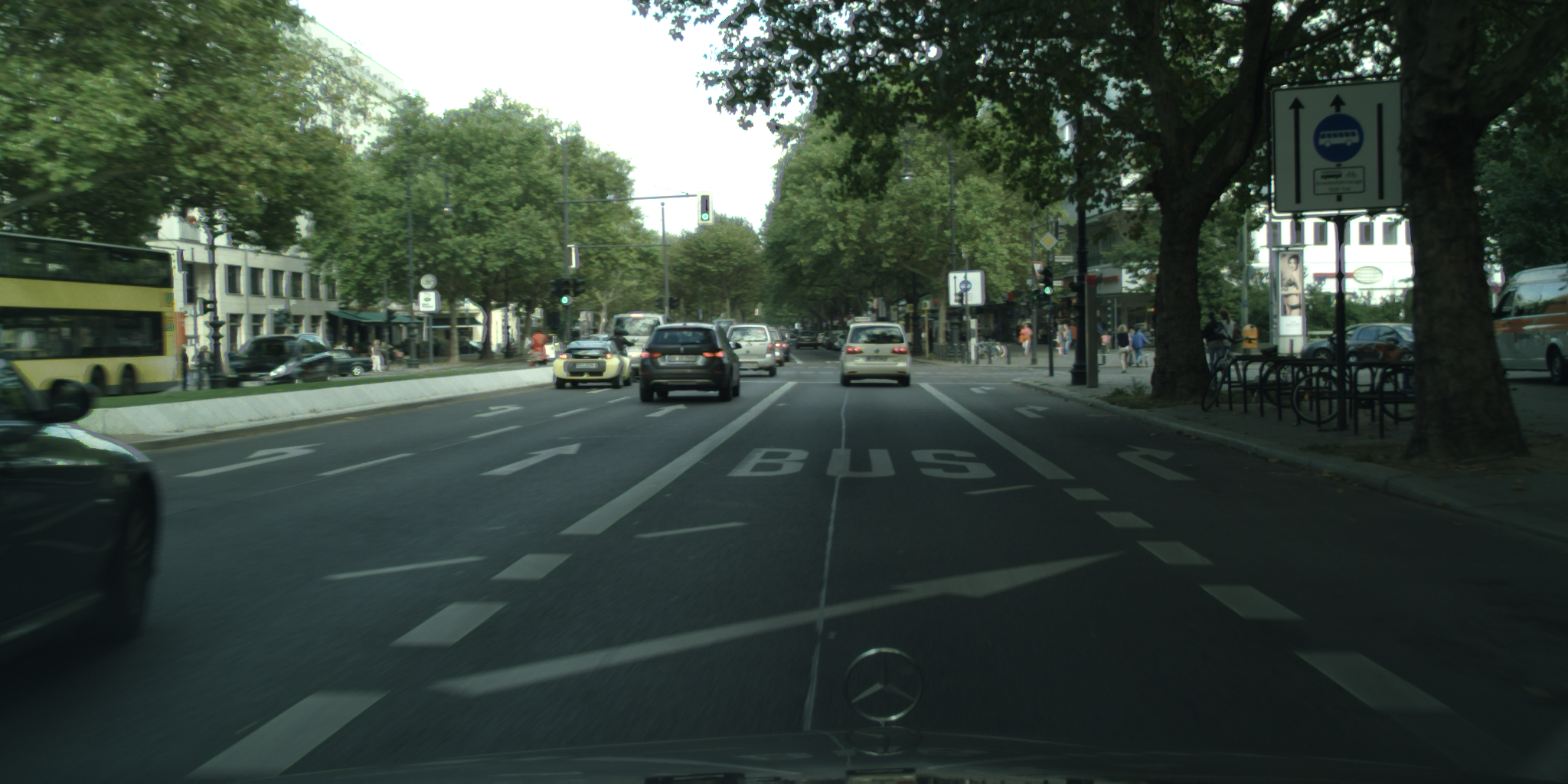}} &
        {\includegraphics[width=0.35\columnwidth]{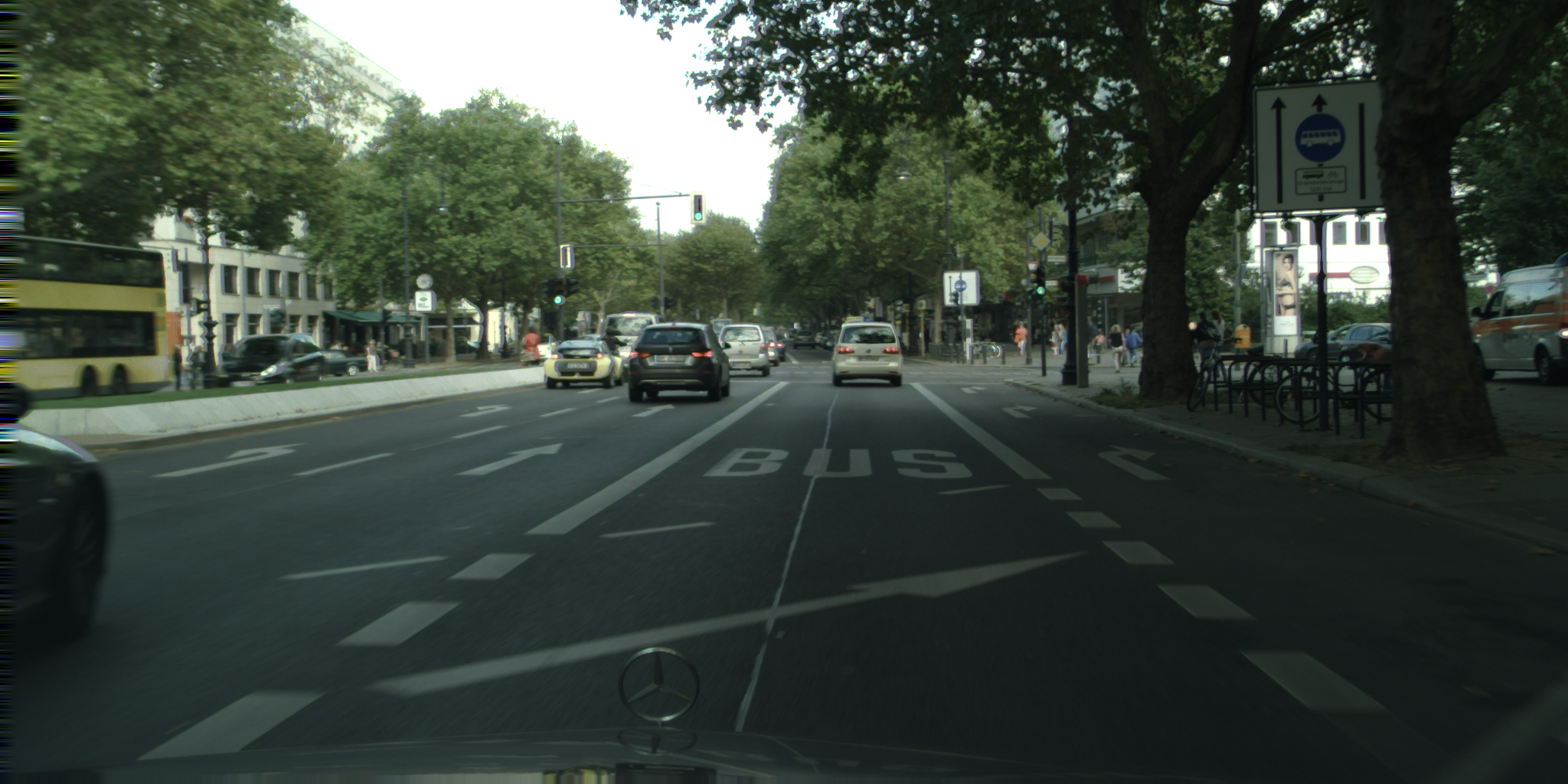}}\\
         (e) & (f)\\
     \end{tabular}
    \caption{Some representative images from the synthetic and real datasets. (a-b) CVGL (c-d) Tsinghua-Daimler (e-f) Cityscapes.}
    \label{fig:foobar}
\end{figure}

\section{Results and Evaluation}



\begin{table*}[t]

\caption{Table showing Normalized MAE in predicted parameters on CVGL test set comprising of 36,905 images.}
\centering
{
\begin{adjustbox}{max width=\textwidth}
\begin{tabular}{ |l|c|c|c|c|c|c|c|c|c|c| } 
\hline
Method & $f_x$ & $f_y$ & $u_0$ & $v_0$ & $b$ & $d$ & $t_x$ & $t_y$ & $t_z$ & $\theta_p$\\
 \hline
Average~\cite{workman2015deepfocal} & 1.003 & 1.539 & 1.326 & 1.200 & \textbf{-1.908} & \textbf{-6.624} & -0.562 & 94.233 & \textbf{-14.031} & \textbf{-3.123}\\
Deep-Homo~\cite{detone2016deep} & 0.055 & 0.055 & 0.018 & 0.018 & -0.091 & -0.167 & -0.091 & 1.900 & -1.55 & -0.258\\
Deep-Calib~\cite{bogdan2018deepcalib} & 0.015 & 0.015 & 0.015 & 0.013 & -0.127 & -0.146 & -0.126 & 0.292 & -0.511 & -0.308\\
MN-CPL-U (Ours) & 0.713 & 0.573 & 0.854 & 0.790 & -0.310 & -1.447 & \textbf{-0.722} & 3.085 & -1.165 & -1.392\\
MN-Baseline (Ours)  & 0.387 & 0.445 & 0.311 & 0.313 & -0.240 & -0.449 & -0.258 & 0.170 & -0.424 & -0.405\\
MN-CPL-A (Ours) & \textbf{0.005} & \textbf{0.003} & 0.007 & 0.011 & -0.013 & -0.121 & -0.013 & 0.093 & -0.050 & -0.024\\
SN-CPL-U (Ours) & 0.134 & 0.140 & 0.127 & 0.068 & -0.274 & -0.345 & -0.596 & 2.241 & -0.820 & -1.672\\
SN-Baseline (Ours)  & 0.007 & 0.009 & \textbf{0.003} & \textbf{0.003} &  -0.011 & -0.117 & -0.014 & \textbf{0.081} & -0.051 & -0.022\\
SN-CPL-A (Ours) & 0.100 & 0.090 & 0.063 & 0.067 & -0.156 & -0.209 & -0.117 & 0.586 & -1.157 & -0.243\\
\hline
\end{tabular}
\end{adjustbox}
\label{table:1}
}
\end{table*}

\begin{table*}[t]

\caption{Table showing Normalized MAE in predicted parameters on Tsinghua-Daimler test set comprising of 2,914 images. For this experiment, we just did a forward pass without any transfer learning or training.}
\centering
{
\begin{adjustbox}{max width=\textwidth}
\begin{tabular}{ |l|c|c|c|c|c|c|c|c|c|c| } 
\hline
 Method& $f_x$ & $f_y$ & $u_0$ & $v_0$ & $b$ & $d$ & $t_x$ & $t_y$ & $t_z$ & $\theta_p$\\
 \hline
Average~\cite{workman2015deepfocal} & 1.000 & 1.014 & 1.017 & 1.021 & 364.633 & 4.893 & 31.167 & 327.128 & 5.109 & 2321.338\\
Deep-Homo~\cite{detone2016deep} & 0.971 & 0.971 & 0.947 & 0.896 & 350.476 & 1.538 & 37.494 & \textbf{10.470} & 2.182 & \textbf{193.760}\\
Deep-Calib~\cite{bogdan2018deepcalib} & 0.962 & 0.962 & 0.947 & 0.895 & \textbf{135.246} & 1.586 & \textbf{15.066} & 23.718 & \textbf{1.379} & 443.233\\
MN-CPL-U (Ours) & \textbf{0.950} & \textbf{0.944} & \textbf{0.896} &  \textbf{0.869} & 173.574 & 2.634 & 81.817 & 44.099 & 1.497 & 290.614\\
MN-Baseline (Ours)  & 0.958 & 0.958 & 0.946 & 0.894 & 237.554 & 1.563 & 26.660 & 27.387 & 1.637 & 354.791\\
MN-CPL-A (Ours) & 0.952 & 0.956 & 0.946 & 0.895 & 363.883 & 1.515 & 44.958 & 33.421 & 1.785 & 324.892\\
SN-CPL-U (Ours) & 0.974 & 0.973 & 0.945 & 0.894 & 168.096 & 1.683 & 32.619 & 17.754 & 1.447 & 287.237\\
SN-Baseline (Ours)  & 0.968 & 0.966 & 0.946 & 0.894 & 276.792 & \textbf{1.507} & 28.885 & 24.962 & 1.606 & 406.246\\
SN-CPL-A (Ours) & 0.966 & 0.966 & 0.946 & 0.894 & 367.383 & 1.514 & 38.998 & 22.145 & 1.487 & 386.369\\
\hline
\end{tabular}
\end{adjustbox}
\label{table:2}
}
\end{table*}

\begin{table*}[t]

\caption{Table showing Normalized MAE in predicted parameters on Cityscapes test set comprising of 1,525 images. For this experiment, we just did a forward pass without any transfer learning or training.}
\centering
{
\begin{adjustbox}{max width=\textwidth}
\begin{tabular}{ |l|c|c|c|c|c|c|c|c|c|c| } 
\hline
 Method& $f_x$ & $f_y$ & $u_0$ & $v_0$ & $b$ & $d$ & $t_x$ & $t_y$ & $t_z$ & $\theta_p$\\
 \hline
Average~\cite{workman2015deepfocal} & 1.000 & 1.014 & 1.016 & 1.021 & 356.941 & 6.659 & 36.490 & 1061.812 & 5.116 & 1225.327\\
Deep-Homo~\cite{detone2016deep} & 0.967 & 0.967 & 0.948 & 0.892 & 438.730 & 1.849 & 55.932 & \textbf{37.543} & 2.338 & \textbf{107.539}\\
Deep-Calib~\cite{bogdan2018deepcalib} & 0.963 & 0.962 & 0.948 & 0.892 & 136.451 & 1.874 & \textbf{18.078} & 75.365 & \textbf{1.385} & 198.051\\
MN-CPL-U (Ours) & 0.961 & 0.958 & \textbf{0.933} & \textbf{0.882} & \textbf{114.965} & 2.521 & 38.778 & 101.388 & 1.500 & 166.183\\
MN-Baseline (Ours)  & 0.960 & 0.960 & 0.947 & 0.891 & 230.870 & 1.884 & 32.058 & 82.282 & 1.843 & 178.436\\
MN-CPL-A (Ours) & \textbf{0.952} & \textbf{0.955} & 0.948 & 0.892 & 312.144 & 1.790 & 50.763 & 112.063 & 1.840 & 126.388\\
SN-CPL-U (Ours) & 0.970 & 0.969 & 0.947 & 0.891 & 167.936 & 2.063 & 36.100 & 66.122 & 1.465 & 152.504\\
SN-Baseline (Ours)  & 0.960 & 0.959 & 0.948 & 0.891 & 236.612 & 1.785 & 30.039 & 98.217 & 1.470 & 178.193\\
SN-CPL-A (Ours) & 0.962 & 0.962 & 0.948 & 0.892 & 194.548 & \textbf{1.783} & 25.318 & 81.614 & 1.490 & 153.042\\
\hline
\end{tabular}
\end{adjustbox}
\label{table:3}
}
\end{table*}



\begin{table*}[t]

\caption{Table showing Accuracy on different thresholds using Horizontal Field of View in predicted parameters on CVGL test set comprising of 36,905 images.}
\centering
{
\begin{adjustbox}{max width=\textwidth}
\begin{tabular}{ |l|c|c|c|c|c|c| } 
\hline
Method & 0 & 1 & 2 & 3 & 4 & 5\\
 \hline
Average~\cite{workman2015deepfocal} & 0.0 & 0.0 & 0.0 & 0.0 & 0.0 & 0.584\\
Deep-Homo~\cite{detone2016deep} & 0.0 & 1.0 & 1.0 & 1.0 & 1.0 & 1.0\\
Deep-Calib~\cite{bogdan2018deepcalib} & 0.0 & 1.0 & 1.0 & 1.0 & 1.0 & 1.0\\
MN-CPL-U (Ours) & 0.0 & 0.976 & 0.989 & 1.0 & 1.0 & 1.0\\
MN-Baseline (Ours)  & 0.0 & 0.983 & 0.991 & 1.0 & 1.0 & 1.0\\
MN-CPL-A (Ours) & 0.0 & \textbf{1.0} & \textbf{1.0} & 1.0 & 1.0 & 1.0\\
SN-CPL-U (Ours) & 0.0 & 0.982 & 1.0 & 1.0 & 1.0 & 1.0\\
SN-Baseline (Ours)  & 0.0 & 0.999 & 1.0 & 1.0 &  1.0 & 1.0\\
SN-CPL-A (Ours) & 0.0 & 0.992 & 1.0 & 1.0 & 1.0 & 1.0\\
\hline
\end{tabular}
\end{adjustbox}
\label{table:hf1}
}
\end{table*}

\begin{table*}[t]

\caption{Table showing Accuracy on different thresholds using Horizontal Field of View in predicted parameters on Tsinghua-Daimler test set comprising of 2,914 images.}
\centering
{
\begin{adjustbox}{max width=\textwidth}
\begin{tabular}{ |l|c|c|c|c|c|c| } 
\hline
Method & 0 & 1 & 2 & 3 & 4 & 5\\
 \hline
Average~\cite{workman2015deepfocal} & 0.0 & 0.0 & 0.0 & 0.0 & 1.0 & 1.0\\
Deep-Homo~\cite{detone2016deep} & 0.0 & 0.019 & \textbf{1.0} & 1.0 & 1.0 & 1.0\\
Deep-Calib~\cite{bogdan2018deepcalib} & 0.0 & 0.303 & 0.982 & 1.0 & 1.0 & 1.0\\
MN-CPL-U (Ours) & 0.0 & 0.127 & 0.867 & 1.0 & 1.0 & 1.0\\
MN-Baseline (Ours)  & 0.0 & 0.413 & 0.996 & 1.0 & 1.0 & 1.0\\
MN-CPL-A (Ours) & 0.0 & \textbf{0.812} & 0.995 & 1.0 & 1.0 & 1.0\\
SN-CPL-U (Ours) & 0.0 & 0.109 & 0.765 & 1.0 & 1.0 & 1.0\\
SN-Baseline (Ours)  & 0.0 & 0.276 & 0.808 & 1.0 &  1.0 & 1.0\\
SN-CPL-A (Ours) & 0.0 & 0.203 & 0.924 & 1.0 & 1.0 & 1.0\\
\hline
\end{tabular}
\end{adjustbox}
\label{table:hf2}
}
\end{table*}

\begin{table*}[t]

\caption{Table showing Accuracy on different thresholds using Horizontal Field of View in predicted parameters on Cityscapes test set comprising of 1,525 images.}
\centering
{
\begin{adjustbox}{max width=\textwidth}
\begin{tabular}{ |l|c|c|c|c|c|c| } 
\hline
Method & 0 & 1 & 2 & 3 & 4 & 5\\
 \hline
Average~\cite{workman2015deepfocal} & 0.0 & 0.0 & 0.0 & 0.0 & 1.0 & 1.0\\
Deep-Homo~\cite{detone2016deep} & 0.0 & 0.046 & \textbf{1.0} & 1.0 & 1.0 & 1.0\\
Deep-Calib~\cite{bogdan2018deepcalib} & 0.0 & 0.253 & 0.992 & 1.0 & 1.0 & 1.0\\
MN-CPL-U (Ours) & 0.0 & 0.046 & 0.933 & 1.0 & 1.0 & 1.0\\
MN-Baseline (Ours)  & 0.0 & 0.330 & 0.996 & 1.0 & 1.0 & 1.0\\
MN-CPL-A (Ours) & 0.0 & \textbf{0.817} & 0.998 & 1.0 & 1.0 & 1.0\\
SN-CPL-U (Ours) & 0.0 & 0.164 & 0.844 & 1.0 & 1.0 & 1.0\\
SN-Baseline (Ours)  & 0.0 & 0.540 & 0.908 & 1.0 &  1.0 & 1.0\\
SN-CPL-A (Ours) & 0.0 & 0.292 & 0.954 & 1.0 & 1.0 & 1.0\\
\hline
\end{tabular}
\end{adjustbox}
\label{table:hf3}
}
\end{table*}


\subsection{Datasets}

SUN360 is the name of the dataset that is frequently used in literature as benchmark data for the estimate of camera parameters~\cite{xiao2012recognizing}. Many of the state-of-the-art methods~\cite{zhang2020deepptz,bogdan2018deepcalib,lopez2019deep} have been evaluated on this dataset however to the best of our knowledge the SUN360 database~\cite{xiao2012recognizing} is no more publicly available. For this reason, we have generated a new dataset using CARLA~\cite{Dosovitskiy17} having 2 towns, sample images along with town maps are shown in Fig.~\ref{fig:foobar} and Fig.~\ref{fig:town} respectively.

\textbf{CVGL Camera Calibration Dataset}: By creating a fresh CVGL Camera Calibration dataset, we trained and assessed~\cite{butt2022camera} using Town 1 and Town 2 of CARLA~\cite{Dosovitskiy17} Simulator. 
The dataset consists of $50$ camera configurations with each town having $25$ configurations. The parameters modified for generating the configurations include  $fov$, $x$, $y$, $z$, pitch, yaw, and roll. Here, $fov$ is the field of view, (x, y, z) is the translation while (pitch, yaw, and roll) is the rotation between the cameras. The total number of image pairs is $1,23,017$, out of which $58,596$ belong to Town 1 while $64,421$ belong to Town 2, the difference in the number of images is due to the length of the tracks.

\textbf{Cyclist Detection Dataset}: Cyclists are divided into three classes as shown in Fig.~\ref{fig:cdbc}. We have used a recent Cyclist Detection dataset~\cite{li2016new} for evaluating our approach on real world data. 
We have used the test set provided by the authors containing $2,914$ images by first deriving the right image using left and disparity images and then use the pair as input to compare different methods.   

\textbf{Cityscapes Dataset}: We have used the Cityscapes dataset~\cite{Cordts2016Cityscapes} for evaluating our approach on real world data. 
We have used the test set provided by the authors containing $1,525$ images by first deriving the right image using left and disparity images and then use the pair as input to compare different methods.   





\textbf{Implementation Details}: Keras is used to implement and train our loss~\cite{ketkar2017introduction}, an open-source deep learning framework. All networks are trained on GeForce GTX $1050$ Ti GPU for $200$ epochs with early stopping using ADAM optimizer~\cite{KingmaADAM2015} with Mean Absolute Error (MAE) loss function and a base learning rate $\eta$ of ${10}^{-3}$ with a batch size of $16$.

When creating Sequential and Functional API models, Keras' Lambda layer enables the use of any expression as a Layer. Lambda layers have been used in the suggested design for fundamental operations like addition, subtraction, multiplication, division, negation, cosine, and sine. It makes sense to include mathematical equations in the framework for learning. Lambda layer representation of $x_{cam}$, $y_{cam}$ and $z_{cam}$ is shown in Fig.~\ref{fig:xcam}, Fig.~\ref{fig:ycam} and Fig.~\ref{fig:zcam} respectively.

There is no universal agreement on how to evaluate calibration networks; some earlier works use a classification approach and provide accuracy values directly~\cite{workman2016horizon}. Others set a cutoff for the regression mistakes and report accuracy values~\cite{hold2018perceptual,workman2015deepfocal,lopez2019deep}. Yin~\emph{et al.} ~\cite{yin2018fisheyerecnet} report peak signal-to-noise ratio structural similarity errors. Rong~\emph{et al.}~\cite{rong2016radial} use a metric based on straight line segment lengths that is only meaningful for radial distortion correction.

We have used Normalized Mean Absolute Error for evaluation as follows:

  \begin{equation}
    \label{eq-60}
    NMAE(y, \hat{y}) = \frac{MAE(y, \hat{y})}{\frac{1}{n}\sum_{i=1}^{n}|y_{i}|} = \frac{MAE(y, \hat{y})}{mean(|y|)}
  \end{equation}
where $y$ and $\hat{y}$ are the target and estimated values respectively.

We have also used Horizontal feild of view as follows:

  \begin{equation}
    \label{eq-60}
    H_{\theta} = 2tan^{-1}(\frac{w}{2f})
  \end{equation}
  
for a given image width, $w$.

\begin{figure}
\centering
    {\includegraphics[trim={0 11cm 6cm 2cm},clip,width=1.0\columnwidth]{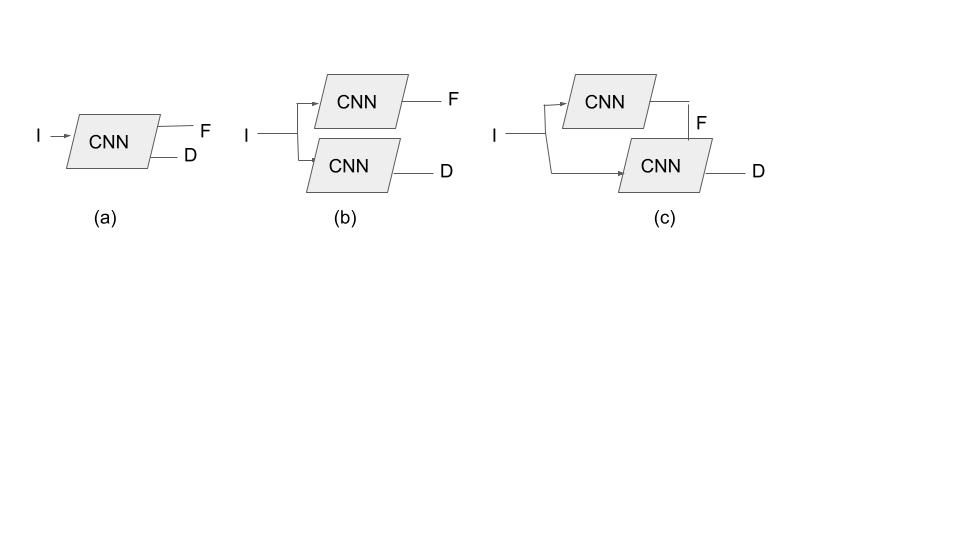}} 
    \caption{Three network architectures suggested by Deep-Calib are shown in an illustration: SingleNet (a), DualNet (b), and SeqNet (c). The calibration image serves as the input, and the distortion parameter and focal length are the outputs. In SeqNet, one of the second network's dense layers is combined with the focal length estimate from the first network.~\cite{bogdan2018deepcalib}}
    \label{fig:dc}
\end{figure}

\subsection{Comparative Analysis}

\textbf{Experimental Setup}: We contrasted our suggested strategy with three cutting-edge techniques, including Average field of view~\cite{workman2015deepfocal}, Deep-Homo~\cite{detone2016deep} and Deep-Calib~\cite{bogdan2018deepcalib}. Average field of view~\cite{workman2015deepfocal} is a standard method that predicts the average field of view of the training set given a query image~\cite{workman2015deepfocal}. {Deep-Homo}~\cite{detone2016deep} estimates the homography of two photos at 8 degrees of freedom. An overview of the approach can be seen in Fig.~\ref{fig:dh}. We have modified Deep-Homo\cite{detone2016deep} must use 4-point parameterization and then convert the result into the homography matrix in order to predict the necessary 13 parameters for comparison purposes as by default, it only predicted 8 values corresponding to the four corners. We have modified Deep-Calib~\cite{bogdan2018deepcalib}, while by default, it only anticipated 2 values for focal length and distortion, it was necessary to forecast the necessary 13 characteristics in order to do comparisons. Overview of the architectures proposed by Deep-Calib~\cite{bogdan2018deepcalib} is shown in Fig.~\ref{fig:dc}. We also developed Baseline, CPL-U, and CPL-A versions of our multi-task learning technique specifically for the ablative study. Baseline is an end-to-end learning architecture based on mean absolute error and does not include any further layers to incorporate camera model equations (MAE). It has 13 regressors that all use the same feature extractor to forecast the necessary values. To investigate the impact of the suggested camera projection loss, baseline is used. In order to balance the disparate ranges of calibration parameters, we additionally used two variations of camera projection loss, one with uniform weighting (CPL-U) in the loss function and the other with adaptive weighting (CPL-A). MN is for Multi-net, where two networks—one for each right and left image—are used, and the features are then combined, whereas SN stands for Single-net, where the combined images are sent to only one network.



\textbf{Error Analysis on CVGL Camera Calibration Dataset}: We contrast the normalised mean absolute error (NMAE) of each parameter using all available techniques with the suggested strategy. It can be seen from Table~\ref{table:1} that for ($f_x$) and ($f_y$) MN-CPL-A performs better while for ($u_0$), ($v_0$)and ($t_y$) , SN-Baseline approach resulted in minimum values for NMAE  and for ($t_x$) MN-CPL-U performs the best due to bias in loss introduced as a result of the heterogeneous range of values among parameters. Overall our method performs better on 6 out of 10 parameters which indicates that incorporating camera model geometry in the learning framework not only resulted in a more interpretable learning framework but it also outperforms the state-of-the-art methods. It can be seen from Table~\ref{table:hf1} that MN-CPL-A performs  best as it has the highest accuracy for thresholds of 1 and 2 as far as horizontal field of view is concerned which means our method outperforms other methods.

\textbf{Error Analysis on Cyclist Detection Dataset}: For this experiment, we didn't trained on the Tsinghua-Daimler dataset but just performed a forward pass to test the generalizability and the results further strengthen our argument. It can be seen from Table~\ref{table:2} that our proposed multi-task learning approach outperforms other methods on $5$ out of $10$ parameters. For disparity ($d$), MN-CPL-U resulted in higher NMAE values due to bias in loss introduced as a result of the heterogeneous range of values among parameters which further solidifies  our argument of incorporating camera model geometry in the learning framework. It can be seen from Table~\ref{table:hf2}  that MN-CPL-A performs for threshold of 1 while Deep-Homo performs best for threshold of 2 which shows our approach is at par compared to other methods

\textbf{Error Analysis on Cityscapes Dataset}: For this experiment, we didn't trained on the Cityscapes dataset but just performed a forward pass to test the generalizability and the results further strengthen our argument. It can be seen from Table~\ref{table:3} that our proposed multi-task learning approach outperforms other methods on $6$ out of $10$ parameters which further solidifies  our argument of incorporating camera model geometry in the learning framework. For disparity ($d$), pitch ($\theta_p$) and  ($t_x$), MN-CPL-U resulted in higher NMAE values due to bias in loss introduced as a result of the heterogeneous range of values among parameters. It  can be seen from Table~\ref{table:hf3} that MN-CPL-A performs best for threshold of 1 while Deep-Homo  works best for  threshold of 2 which suggests  that our approach is at par compared to other methods.

\section{Conclusion}
On both synthetic and actual data, our suggested strategy surpasses numerous benchmarks, including CNN-based techniques. It combines Deep Learning and closed-form analytical approaches. 
As a result, this work presents a fundamentally novel concept in the fields of machine learning and computer vision. Despite the fact that we used the suggested technique to estimate camera calibration parameters, it provides a basic framework for many similar issues. For instance, Kalman Filter tracking, homography estimation, and many other applications are among those we want to use in the future.


\bibliographystyle{IEEEbib}
\bibliography{strings,refs}

\end{document}